\def\year{2020}\relax
\documentclass[letterpaper]{article} 
\usepackage{aaai20}  
\usepackage{times}  
\usepackage{helvet}  
\usepackage{courier}  
\usepackage{url}  
\usepackage{graphicx}  
\usepackage{amsmath}
\usepackage{amssymb}
\usepackage{pifont}
\usepackage{soul}

\nocopyright

\usepackage{enumitem}
\usepackage{tabularx,tabulary}
\usepackage{algorithm2e}
\usepackage{algpseudocode}
\usepackage{amsmath,amssymb} 
\usepackage[table]{xcolor} 

\definecolor{coolblack}{rgb}{0.0, 0.23, 0.64}
\usepackage[pagebackref=true,breaklinks=true,letterpaper=true,colorlinks,bookmarks=false]{hyperref}
\hypersetup{
     citecolor = coolblack,
}

\usepackage{multirow}
\usepackage{makecell}
\usepackage{amsmath}
\usepackage{bm}

\DeclareRobustCommand{\uvec}[1]{{%
  \ifcsname uvec#1\endcsname
     \csname uvec#1\endcsname
   \else
    \bm{\hat{\mathbf{#1}}}%
   \fi
}}

\usepackage{array}
\newcolumntype{L}[1]{>{\raggedright\let\newline\\\arraybackslash\hspace{0pt}}m{#1}}
\newcolumntype{C}[1]{>{\centering\let\newline\\\arraybackslash\hspace{0pt}}m{#1}}
\newcolumntype{R}[1]{>{\raggedleft\let\newline\\\arraybackslash\hspace{0pt}}m{#1}}

\usepackage{xspace}
\makeatletter
\DeclareRobustCommand\onedot{\futurelet\@let@token\@onedot}
\def\@onedot{\ifx\@let@token.\else.\null\fi\xspace}
\def\eg{\emph{e.g}\onedot} 
\def\ie{\emph{i.e}\onedot}

\makeatother

\newcommand{\citet}[1]{\citeauthor{#1} \shortcite{#1}}
\newcommand{\citep}{\cite}

\definecolor{dgreen}{rgb}{0.04,0.7,0.13}
\newcommand{\cmark}{{\color{red}\ding{51}}}%
\newcommand{\xmark}{{\color{dgreen}\ding{55}}}%

\begin{document}
%
\title{Kinematic-Structure-Preserved Representation for \\Unsupervised 3D Human Pose Estimation}
\author{Jogendra Nath Kundu\thanks{equal contribution}\qquad Siddharth Seth\footnotemark[1]\qquad Rahul M V\footnotemark[1]\qquad Mugalodi Rakesh\qquad \\
\bf \Large R. Venkatesh Babu\qquad Anirban Chakraborty
\\
Indian Institute of Science, Bangalore, India\\
{\small \{jogendrak, siddharthseth\}@iisc.ac.in, rmvenkat@andrew.cmu.edu, rakeshramesha@gmail.com, \{venky, anirban\}@iisc.ac.in}}

\maketitle
\begin{abstract}
Estimation of 3D human pose from monocular image has gained considerable attention, as a key step to several human-centric applications. However, generalizability of human pose estimation models developed using supervision on large-scale in-studio datasets remains questionable, as these models often perform unsatisfactorily on unseen in-the-wild environments. Though weakly-supervised models have been proposed to address this shortcoming, performance of such models relies on availability of paired supervision on some related tasks, such as 2D pose or multi-view image pairs. In contrast, we propose a novel kinematic-structure-preserved unsupervised 3D pose estimation framework\footnote{{\url{https://sites.google.com/view/ksp-human/}}}, which is not restrained by any paired or unpaired weak supervisions. Our pose estimation framework relies on a minimal set of prior knowledge that defines the underlying kinematic 3D structure, such as skeletal joint connectivity information with bone-length ratios in a fixed canonical scale. The proposed model employs three consecutive differentiable transformations named as forward-kinematics, camera-projection and spatial-map transformation. This design not only acts as a suitable bottleneck stimulating effective pose disentanglement, but also yields interpretable latent pose representations avoiding training of an explicit latent embedding to pose mapper.  Furthermore, devoid of unstable adversarial setup, we re-utilize the decoder to formalize an energy-based loss, which enables us to learn from in-the-wild videos, beyond laboratory settings. Comprehensive experiments demonstrate our state-of-the-art unsupervised and weakly-supervised pose estimation performance on both Human3.6M and MPI-INF-3DHP datasets. Qualitative results on unseen environments further establish our superior generalization ability.

\end{abstract}

\begin{table}[t]
	\footnotesize
	\caption{  
	Characteristic comparison of our approach against prior unsupervised and weakly-supervised human 3D pose estimation works, in terms of access to direct (paired) or indirect (unpaired) supervision levels (MV: Multi-View). Note that, in the proposed framework \textbf{the latent pose representation itself, is the 3D pose coordinates}, thereby avoiding training of a separate latent to 3D pose mapper (last column).}
	\centering
	\setlength\tabcolsep{2.4pt}
	\resizebox{0.47\textwidth}{!}{
	\begin{tabular}{l|ccc|cc|c}
	\hline
 		\multirow{2}{*}{Methods} & \multicolumn{3}{c|}{Paired sup.} &
 		\multicolumn{2}{c|}{\makecell{Unpaired sup.\\(adv. learning)}} & \multirow{2}{*}{\makecell{\vspace{-5.5mm}\\Sup. for \\latent to\\3D pose\\ mapping}} \\
 		\cline{2-6} 
 		& \makecell{2D \\pose} & \makecell{MV\\ pair} & \makecell{Cam.\\ extrin.} & \makecell{2D \\pose} & \makecell{3D \\pose} \\ \hline\hline
 		\rowcolor{gray!00}
 		({\color{coolblack}Rhodin et al.} \citeyear{rhodin2018unsupervised})&\xmark &\cmark &\cmark &\xmark &\xmark &\cmark 
 		\\
 		\rowcolor{gray!10}
 		({\color{coolblack}kocabas et al.} \citeyear{kocabas2019self}) &\cmark &\cmark &\xmark &\xmark &\xmark &\xmark     \\
 		\rowcolor{gray!00}
 		\cite{chen2019weakly} &\cmark &\cmark &\xmark &\xmark &\xmark &\cmark 
 		\\
 		\rowcolor{gray!10}
 		({\color{coolblack}Wandt et al.} \citeyear{wandt2019repnet}) &\cmark &\xmark &\xmark &\xmark &\cmark &\xmark    \\
 		\rowcolor{gray!00}
 		\cite{chen2019unsupervised} &\cmark &\xmark &\xmark &\cmark &\xmark &\xmark    \\
 		\rowcolor{gray!20}
 		Ours (unsup.) &\xmark &\xmark &\xmark &\xmark &\xmark &\xmark   \\ 
		\hline
	\end{tabular}}
	\vspace{-2mm}
	\label{tab:char}
\end{table} 

\section{Introduction}
Building general intelligent systems, capable of understanding the inherent 3D structure and pose of non-rigid humans from monocular RGB images, remains an illusive goal in the vision community.
In recent years, researchers aim to solve this problem by leveraging the advances in two key aspects, \ie a) improved architecture design~\cite{newell2016stacked,chu2017multi} and b) increasing collection of diverse annotated samples to fuel the supervised learning paradigm~\cite{VNect_SIGGRAPH2017}.
However, obtaining 3D pose ground-truth for non-rigid human-bodies is a highly inconvenient process. Available motion capture systems, such as body-worn sensors (IMUs) or multi-camera structure-from-motion (SFM), requires careful pre-calibration, and hence usually done in a pre-setup laboratory environment~\cite{ionescu2013human3,zhang2017martial}. 
This often restricts diversity in the collected dataset, which in turn hampers generalization of the supervised models trained on such data. For instance, the widely used Human3.6M~\cite{ionescu2013human3} dataset captures 3D pose using 4 fixed cameras (\ie only 4 backgrounds scenes), 11 actors (\ie limited apparel variations), and 17 action categories (\ie limited pose diversity). A model trained on this dataset delivers impressive results when tested on samples from the same dataset, but does not generalize to an unknown deployed environment, thereby yielding non-transferability issue. 

To deal with this problem, researchers have started exploring innovative techniques to reduce dependency on annotated real samples. Aiming to enhance appearance diversity on known 3D pose samples (CMU-MoCap), synthetic datasets have been proposed, by compositing a diverse set of human template foregrounds with random backgrounds~\cite{varol2017learning}. However, models trained on such samples do not generalize to a new motion (e.g. a particular dance form), apparel, or environment much different from the training samples, as a result of large domain shift. Following a different direction, several recent works propose weakly-supervised approaches~\cite{zhou2017towards}, where they consider access to a large-scale dataset with paired supervision on some related-tasks other than the task in focus (\ie 3D pose estimation). Particularly, they access multiple cues for weak supervision, such as, a) paired 2D ground-truth, b) unpaired 3D ground-truth (3D pose without the corresponding image), c) multi-view image pair~({\color{coolblack}Rhodin et al.~\citeyear{rhodin2018unsupervised}}), d) camera parameters in a multi-view setup etc. (see Table~\ref{tab:char} for a detailed analysis).

While accessing such weak paired-supervisions, the general approach is to formalize a self-supervised consistency loop, such as 2D$\rightarrow$3D$\rightarrow$2D~\cite{tung2017adversarial}, view-1$\rightarrow$3D$\rightarrow$view-2~\cite{kocabas2019self}, etc. However, the limitations of domain-shift still persists as a result of using annotated data (\eg 2D ground-truth or multi-view camera extrinsic). To this end, without accessing such paired samples,~\cite{jakab2019learning} proposed to leverage unpaired samples to model the natural distribution of the expected representations (\ie 2D or 3D pose) using adversarial learning. Obtaining such samples, however, requires access to a 2D or 3D pose dataset and hence the learning process is still biased towards the action categories presented in that dataset. One can not expect to have access to any of the above discussed paired or unpaired weak supervisory signals for an unknown deployed environment (e.g. frames of a dance-show where the actor is wearing a rare traditional costume). This motivates us to formalize a fully-unsupervised framework for monocular 3D pose estimation, where the pose representation can be adapted to the deployed environment by accessing only the RGB video frames devoid of dependency on any explicit supervisory signal.

\textbf{Our contributions.} We propose a novel unsupervised 3D pose estimation framework, relying on a carefully designed kinematic structure preservation pipeline. Here, we constrain the latent pose embedding, to form interpretable 3D pose representation, thus avoiding the need for an explicit latent to 3D pose mapper. 
Several recent approaches aim to learn a prior characterizing kinematically plausible 3D human poses using available MoCap datasets~({\color{coolblack}Kundu et al. }\citeyear{kundu2019bihmp}). In contrast, we plan to utilize minimal kinematic prior information, by adhering to the restrictions to not use any external unpaired supervision. This involves, a) access to the knowledge of hierarchical limb connectivity, b) a vector of allowed bone length ratios, and c) a set of 20 synthetically rendered images with diverse background and pose (\ie a minimal dataset with paired supervision to standardize the model towards the intended 2D or 3D pose conventions). The aforementioned prior information is very minimal in comparison to the pose-conditioned limits formalized by~({\color{coolblack}Akhter et al. }\citeyear{akhter2015pose}) in terms of both dataset size and parameters associated to define the constraints.

In the absence of multi-view or depth information, we infer 3D structure, directly from the video samples, for the unsupervised 3D pose estimation task. 
One can easily segment moving objects from a video, in absence of any background (BG) motion. However, this is only applicable to in-studio static camera feeds. Aiming to work on in-the-wild YouTube videos 
, we formalize separate unsupervised learning schemes for videos with both static and dynamic BG. In absence of background motion, we form pairs of video frames with a rough estimate of the corresponding BG image, following a training scheme to disentangle foreground-apparel and the associated 3D pose. However, in the presence of BG motion, we lack in forming such consistent pairs, and thus devise a novel energy-based loss on the disentangled pose and appearance representations. In summary,

\begin{itemize}
\item We formalize a novel collection of three differentiable transformations, which not only acts as a bottleneck stimulating effective pose disentanglement but also yields interpretable latent pose representations avoiding training of an explicit latent-to-pose mapper.

\item The proposed energy-based loss, not only enables us to learn from in-the-wild videos, but also improves generalizability of the model as a result of training on diverse scenarios, without ignoring any individual image sample. 


\item We demonstrate \textit{state-of-the-art} unsupervised and weakly-supervised 3D pose estimation performance on both Human3.6M and MPI-INF-3DHP datasets.

\end{itemize}

\section{Related Works}
\label{sec:related-works}
\textbf{3D human pose estimation.}
There is a plethora of fully-supervised 3D pose estimations works~\cite{fang2018learning,mehta2017monocular,VNect_SIGGRAPH2017}, where the performance is bench-marked on the same dataset, which is used for training. Such approaches do not generalize on minimal domain shifts beyond the laboratory environment. In absence of large-scale diverse outdoor datasets with 3D pose annotations, datasets with 2D pose annotations is used as a weak supervisory signal for transfer learning using various 2D to 3D lifting techniques ({\color{coolblack}Tung et al.~\citeyear{tung2017adversarial}}; {\color{coolblack}Chen et al.~\citeyear{chen20173d}}; {\color{coolblack}Ramakrishna et al.~\citeyear{ramakrishna2012}}). However, these approaches still rely on availability of 2D pose annotations. Avoiding this, ({\color{coolblack}Kocabas et al.~\citeyear{kocabas2019self}}; {\color{coolblack}Rhodin et al.~\citeyear{rhodin2018unsupervised}}) proposed to use multi-view correspondence acquired by synchronized cameras. But in such approaches ({\color{coolblack}Rhodin et al.~\citeyear{rhodin2018unsupervised}}), the latent pose representation remains un-interpretable and abstract, thereby requiring a substantially large amount of 3D supervision to explicitly train a \textit{latent-to-pose mapping} mapper. We avoid training of such explicit mapping, by casting the latent representation, itself as the 3D pose coordinates. This is realized as a result of formalizing the geometry-aware bottleneck.

\textbf{Geometry-aware representations.}
To capture intrinsic structure of objects, the general approach is to disentangle individual factors of variations, such as appearance, camera viewpoint and other pose related cues, by leveraging inter-instance correspondence. In literature, we find unsupervised land-mark detection techniques~\cite{zhang2018unsupervised}, that aim to utilize a relative transformation between a pair of instances of the same object, targeting the 2D pose estimation task. To obtain such pairs, these approaches rely on either of the following two directions, viz. a) frames from a video with an acceptable time-difference~\cite{jakab2018unsupervised}, or b) synthetically simulated 2D transformations~\cite{rocco2017convolutional}. However, such techniques fail to capture the 3D structure of the object in the absence of multi-view information. The problem becomes more challenging for deformable 3D skeletal structures as found in diverse human poses. Recently~\cite{jakab2018unsupervised} proposed an unsupervised 2D landmark estimation method to disentangle pose from appearance using a conditional image generation framework. However, the predicted 2D landmarks do not match with the standard human pose key-points, hence are highly un-interpretable with some landmarks even lying on the background. Such outputs can not be used for a consequent task requiring a structurally consistent 2D pose input. 

Defining structural constraints in 2D is highly ill-posed, considering images as projections of the actual 3D world. Acknowledging this, we plan to estimate 3D pose separately with camera parameters followed by a camera-projection to obtain the 2D landmarks. As a result of this inverse-graphics formalization, we have the liberty to impose structural constraints directly on the 3D skeletal representation, where the bone-length and other kinematic constraints can be imposed seamlessly using consistent rules as compared to the corresponding 2D representation. A careful realization of 3D structural constraints not only helps us to obtain interpretable 2D landmarks but also reduces the inherent uncertainty associated with the process of lifting a monocular 2D images to 3D pose~\cite{chen2019unsupervised}, in absence of any additional supervision such as multi-view or depth cues.

\begin{figure*}
\begin{center}
	\includegraphics[width=1.0\linewidth]{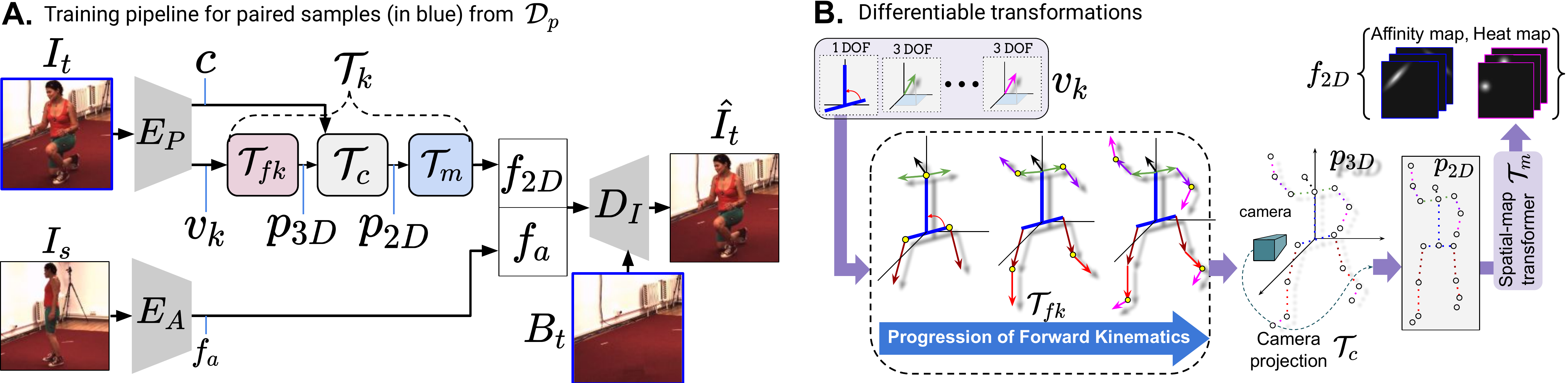}
	\caption{\textbf{A.} Illustration of the proposed framework indicating output notation of individual modules. \textbf{B.} An overview of the three differentiable transformations, with step-wise progression of forward kinematics using local-kinematic parameters, $v_k$.
	}
 	\label{fig:main1}    
    \vspace{-4mm}
\end{center}
\end{figure*}


\section{Approach}
\label{sec:approach}
Our aim is to learn a mapping function, that can map an RGB image of human to its 3D pose by accessing minimal kinematic prior information. Motivated by ({\color{coolblack}Rhodin et al.~\citeyear{rhodin2018unsupervised}}), we plan to cast it as an unsupervised disentanglement of three different factors \ie, a) foreground (FG) appearance, b) background (BG) appearance, and c) kinematic pose. However, unlike~({\color{coolblack}Rhodin et al.~\citeyear{rhodin2018unsupervised}}) in absence of multi-view pairs, we have access to simple monocular video streams of human actions consisting of both static and dynamic BG.

\subsection{Architecture}
As shown in Fig.~\ref{fig:main1}A, we employ two encoder networks each with a different architecture, $E_P$ and $E_A$ to extract the local-kinematic parameters $v_k$ (see below) and FG-appearance, $f_a$ respectively from a given RGB image. Additionally, $E_P$ also outputs 6 camera parameters, denoted by $c$, to obtain coordinates of the camera-projected 2D landmarks, $p_{2D}$.

One of the major challenges in learning factorized representations~\cite{denton2017unsupervised} is to realize purity among the representations. More concretely, the appearance representation should not embed any pose related information and vice-versa. To achieve this, we enforce a bottleneck on the pose representation by imposing kinematic-structure based constraints (in 3D) followed by an inverse-graphics formalization for 3D to 2D re-projection. This introduces three pre-defined transformations \ie, a) Forward kinematic transformation, $\mathcal{T}_{fk}$ and b) Camera projection transformation $\mathcal{T}_c$, and c) Spatial-map transformation $\mathcal{T}_m$. 

\subsubsection{a) Forward kinematic transformation, $\mathcal{T}_{fk}$} Most of the prior 3D pose estimation approaches~({\color{coolblack}Chen et al.~\citeyear{chen2019unsupervised}}; {\color{coolblack}Rhodin et al.~\citeyear{rhodin2018unsupervised}}) aim to either directly regress joint locations in 3D or depth associated with the available 2D landmarks. Such approaches do not guarantee validity of the kinematic structure, thus requiring additional loss terms in the optimization pipeline to explicitly impose kinematic constraints such as bone-length and limb-connectivity information~\cite{habibie2019wild}. In contrast, we formalize a view-invariant local-kinematic representation of the 3D skeleton based on the knowledge of skeleton joint connectivity. 
We define a canonical rule (see Fig.~\ref{fig:main1}B), by fixing the neck and pelvis joint (along z-axis, with pelvis at the origin) and restricting the {trunk to hip-line (line segment connecting the two hip joints) angle}, to rotate only about x-axis on the YZ-plane(\ie 1-DOF) in the canonical coordinate system $C$ (\ie Cartesian system defined at the pelvis as origin). Our network regresses one pelvis to hip-line angle and 13 unit-vectors (all 3-DOF), which are defined at their respective parent-relative local coordinate systems, $L^{Pa(j)}$, where $Pa(j)$ denotes the parent joint of $j$ in the skeletal kinematic tree. Thus, $v_k\in \mathbb{R}^{40}$ (\ie 1+13*3). These predictions are then passed on to the forward-kinematic transformation to obtain the 3D joint coordinates $p_{3D}$ in $C$, \ie $\mathcal{T}_{fk}:v_k\rightarrow p_{3D}$ where $p_{3D}\in \mathbb{R}^{3J}$, with $J$ being the total number of skeleton joints. First, positions of the 3 root joints, $p_{3D}^{(j)}$ for $j$ as left-hip, right-hip and neck, are obtained using the above defined canonical rule after applying the estimate of the {trunk to hip-line angle}, $v_k^{(0)}$.
Let $\textit{len}^{(j)}$ store the length of the line-segment (in a fixed canonical unit) connecting a joint $j$ with $Pa(j)$. Then, $p_{3D}^{(j)}$ for rest of the joints is realized using the following recursive equation, $p_{3D}^{(j)} = p_{3D}^{(Pa(j))}+\textit{len}^{(j)}v_k^{(j)}$. See Fig.~\ref{fig:main1}B (dotted box) for a more clear picture.


\subsubsection{b) Camera-projection transformation, $\mathcal{T}_{c}$} As $p_{3D}$ is designed to be view-invariant, we rely on estimates of the camera extrinsics $c$ (3 angles, each predicted as 2 parameters, the $\sin$ and $\cos$ component), which is used to rotate and translate the camera in the canonical coordinate system $C$, to obtain 2D landmarks of the skeleton (\ie using the rotation and translation matrices, $R_c$ and $T_c$ respectively). Note that, these 2D landmarks are expected to register with the corresponding joint locations in the input image. Thus, the 2D landmarks are obtained as, $p_{2D}^{(j)} = P(R_c*p_{3D}^{(j)}+T_c)$, where $P$ denotes a fixed perspective camera transformation.

\subsubsection{c) Spatial-map transformation, $\mathcal{T}_{m}$} 
\begin{figure*}
\begin{center}
	\includegraphics[width=1.0\linewidth]{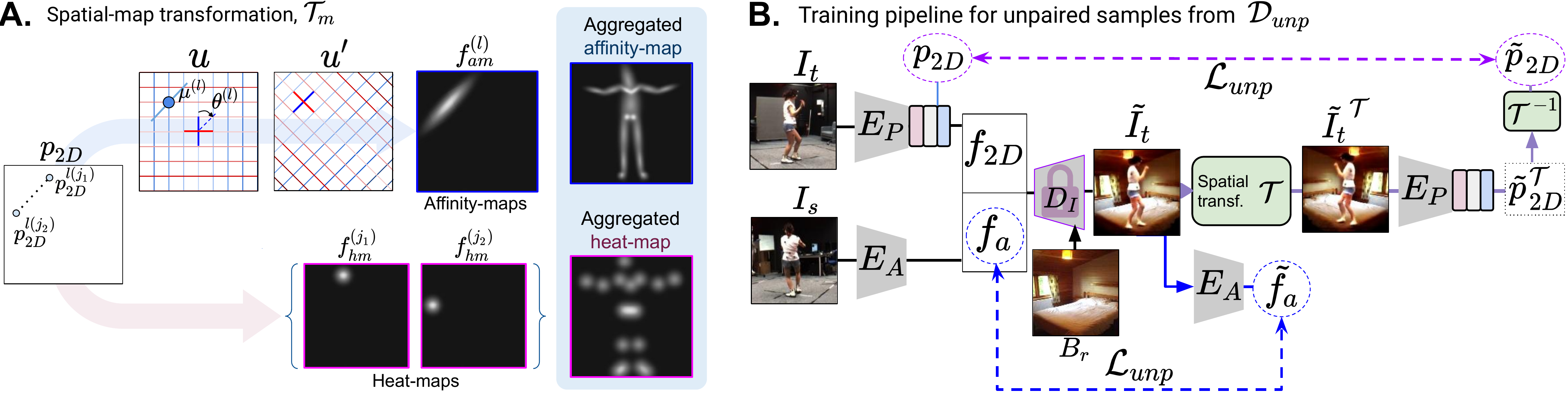}
	\caption{\textbf{A.} Illustration of the steps to obtain the spatial heat-map and affinity-map from the projected 2D coordinates. \textbf{B.} An overview of the proposed data-flow pipeline enabling energy-based loss formalization targeting unpaired samples from $\mathcal{D}_{unp}$.
	}
 	\label{fig:main2}    
    \vspace{-4mm}
\end{center}
\end{figure*}

After obtaining coordinates of the 2D landmarks $p_{2D}\in\mathbb{R}^{2J}$, we aim to effectively aggregate it with the spatial appearance-embedding $f_a$. Thus, we devise a transformation procedure $\mathcal{T}_{m}$, to transform the vectorized 2D coordinates into spatial-maps denoted by $f_{2D}\in \mathbb{R}^{H\times W\times \textit{Ch}}$, which are of consistent resolution to $f_a$, \ie $\mathcal{T}_m:p_{2D}\rightarrow f_{2D}$.
To effectively encode both joint locations and their connectivity information, we propose to generate two sets of spatial maps namely, a) heat-map, $f_{hm}$ and b) affinity-map, $f_{am}$ (\ie, $f_{2D}:(f_{hm},f_{am})$). Note that, the transformations to obtain these spatial maps must be fully differentiable to allow the disentaglement of pose using the cross-pose image-reconstruction loss
, computed at the decoder output (discussed in Sec. {\color{red}3.3a}). Keeping this in mind, we implement a novel computational pipeline by formalizing translated and rotated Gaussians to represent both joint positions (\ie $f_{hm}$) and skeleton-limb connectivity (\ie $f_{am}$). We use a constant variance $\sigma$ along both spatial directions to realize the heat-maps for each joint $j$, as $f_{hm}^{(j)}(u) = \exp(-0.5||u-p_{2d}^{(j)}||^2/\sigma^{2})$, where $u:[u_x,u_y]$ 
denotes the spatial-index in a $H\times W$ lattice (see Fig.~\ref{fig:main2}A). 

We formalize the following steps to obtain the affinity maps based on the connectivity of joints in the skeletal kinematic tree (see Fig.~\ref{fig:main2}A). For each limb (line-segment), $l$ with endpoints $p_{2D}^{l(j_1)}$ and $p_{2D}^{l(j_2)}$, we first compute location of its mid-point, $\mu^{(l)}:[\mu_x^{(l)},\mu_y^{(l)}]$ and slope $\theta^{(l)}$. Following this, we perform an affine transformation to obtain, $u^\prime = R_{\theta^{(l)}}*(u-\mu^{(l)})$, where $R_{\theta^{(l)}}$ is the 2D rotation matrix. Let, $\sigma_x^{(l)}$ and $\sigma_y^{(l)}$ denote variance of a Gaussian along both spatial directions representing the limb $l$. We fix $\sigma_y^{(l)}$ from prior knowledge of the limb width. Whereas, $\sigma_x^{(l)}$ is computed as $\alpha*len(l)$ in the 2D euclidean space (see Supplementary). Finally, the affinity map is obtained as,
\begin{eqnarray*}
f_{am}^{(l)}(u) = \exp(-0.5||u_x^\prime/\sigma_x^{(l)}||^2-0.5||u_y^\prime/\sigma_y^{(l)}||^2) 
\end{eqnarray*}

$\mathcal{T}_{fk}$, $\mathcal{T}_{c}$ and $\mathcal{T}_{m}$ (collectively denoted as $\mathcal{T}_k$) are designed using perfectly differentiable operations, thus allowing back-propagation of gradients from the loss functions defined at the decoder output. As shown in Fig.~\ref{fig:main1}A, the decoder takes in a tuple of spatial-pose-map representation and appearance ($f_{2D}$ and $f_a$ respectively, concatenated along the channel dimension) to reconstruct an RGB image. To effectively disentangle BG information in $f_a$, we fuse the background image $B_t$ towards the end of decoder architecture, inline with~({\color{coolblack}Rhodin et al.~\citeyear{rhodin2018unsupervised}}).

\subsection{Access to minimal prior knowledge}
One of the key objectives of this work is to solve the unsupervised pose estimation problem with minimal access to prior knowledge whose acquisition often requires manual annotation or a data collection setup, such as CMU-MoCap
. Adhering to this, we restrict the proposed framework from accessing any paired or unpaired data samples as shown in Table~\ref{tab:char}. Here, we list the specific prior information that has been considered in the proposed framework,
\begin{itemize}[leftmargin=0.35cm]
\item Kinematic skeletal structure (\ie the joint connectivity information) with bone-length ratios in a fixed canonical scale. Note that, {we do not consider access to the kinematic angle limits} for the limb joints, as such angles are highly pose dependent particularly for diverse human skeleton structures~\cite{akhter2015pose}.

\item A set of 20 synthetically rendered SMPL models with diverse 3D poses and FG appearance~\cite{varol2017learning}. We have direct paired supervision loss (denoted by $\mathcal{L}_{prior}$) on these samples to standardize the model towards the intended 2D or 3D pose conventions (see Supplementary).
\end{itemize}

\subsection{Unsupervised training procedure}
In contrast to~\cite{jakab2018unsupervised}, we aim to disentangle foreground (FG) and background (BG) appearances, along with the disentanglement of pose. 
In a generalized setup, we also aim to learn from in-the-wild YouTube videos in contrast to in-studio datasets, avoiding dataset-bias. 

\subsubsection{Separating paired and unpaired samples.}

For an efficient disentanglement, we aim to form image tuples of the form $(I_s,I_t, B_t)$. Here, $I_s$ and $I_t$ are video frames, which have identical FG-appearance with a nonidentical \textit{kinematic-pose} (pairs formed between frames beyond a certain time-difference). As each video-clip captures action of an individual in a certain apparel, \textit{FG-appearance} remains identical among frames from the same video. Here, $B_t$ denotes an estimate of BG image without the human subject corresponding to the image $I_t$, which is obtained as the median of pixel intensities across a time-window including the frame $I_t$. However, such an estimate of $B_t$ is possible only for scenarios with no camera movement beyond a certain time window to capture enough background evidence (\ie static background with a moving human subject).

Given an in-the-wild dataset of videos, we classify temporal clips of a certain duration ($>$5 seconds) into two groups based on the amount of BG motion in that clip. This is obtained by measuring the pixel-wise L2 loss among the frames in a clip, considering human action covers only 10-20\% of pixels in the full video frame (see Supplementary). Following this, we realize two disjoint datasets denoted by $\mathcal{D}_{p}=\{(I_s^{(i)},I_t^{(i)},B_t^{(i)})\}_{i=1}^{N}$ and $\mathcal{D}_{unp}=\{(I_s^{(k)}, I_t^{(k)})\}_{k=1}^M$ as sets of tuples with extractable BG pair (paired) and un-extractable BG pair (unpaired), respectively.

\subsubsection{a) Training objective for paired samples, $\mathcal{D}_p$} As shown in Fig.~\ref{fig:main1}A, given a source and target image (\ie $I_s$ and $I_t$), we aim to transfer the pose of $I_t$ (\ie $f_{2D}$) to the FG-appearance extracted from $I_s$ (\ie $f_a$) and background from $B_t$ to reconstruct $\hat{I}_t$.  Here, the FG and BG appearance information can not leak through pose representation because of the low dimensional bottleneck \ie $p_{2D}\in\mathbb{R}^{2J}$. Moreover, consecutive predefined matrix  and  spatial-transformation  operations  further restrict the framework from leaking appearance information through the pose branch even as low-magnitude signals. Note that, the BG of $I_s$ may not register with the BG of $I_t$, when the person moves in the 3D world (even in a fixed camera scenario) as these images are outputs of an off-the shelf person-detector. As a result of this BG disparity and explicit presence of the clean spatially-registered background $B_t$, $D_I$ catches the BG information directly from $B_t$, thereby forcing $f_a$ to solely model FG-appearance from the apparel-consistent source, $I_s$. Besides this, we also expect to maintain perceptual consistency between $I_t$ and $\hat{I}_t$ through the encoder networks, keeping in mind the later energy-based formalization (next section). Thus, all the network parameters are optimized for the paired samples using the following loss function, $\mathcal{L}_P = |I_t - \hat{I}_t| + \lambda_1|p_{2D}-\hat{p}_{2D}|+\lambda_2|f_a - \hat{f}_a|$. Here, $\hat{p}_{2D} = \mathcal{T}_{k}\circ E_P(\hat{I_t})$ and $\hat{f_a} = E_A(\hat{I_t})$.

\subsubsection{b) Training objective for unpaired samples, $\mathcal{D}_{unp}$} Although, we find a good amount of YouTube videos where human motion (e.g. dance videos) is captured on a tripod mounted static camera, such videos are mostly limited to indoor environments. However, a diverse set of human actions are captured in outdoor settings (e.g. sports related activities), which usually involves camera motion or dynamic BG. Aiming to learn a general pose representation, instead of ignoring the frames from video-clips with dynamic BG, we plan to formalize a novel direction to adapt the parameters of $E_P$ and $E_A$ even for such diverse scenarios.

We hypothesize that the decoder $D_I$ expects the pose and FG-appearance representation in a particular form, satisfying the corresponding input distributions, $P(f_{2D})$ and $P(f_a)$. Here, a reliable estimate of $P(f_{2D})$ and $P(f_a)$ can be achieved solely on samples from $\mathcal{D}_p$ in presence of paired supervision, avoiding \textit{mode-collapse}. More concretely, the parameters of $D_I$ should not be optimized on samples from $\mathcal{D}_{unp}$ (as shown in Fig.~\ref{fig:main2}B with a lock sign). Following this, one can treat $D_I$ analogous to a \textit{critic}, which outputs a reliable prediction (an image of human with pose from $I_t$, FG-appearance from $I_s$ and BG from $B_t$) only when its inputs $f_{2D}$ and $f_{a}$ satisfy the expected distributions- $P(f_{2D})$ and $P(f_a)$ respectively. We plan to leverage this analogy to effectively use the frozen $D_I$ network as an energy function to realize simultaneous adaptation of $E_P$ and $E_A$ for the unpaired samples from $\mathcal{D}_{unp}$.

We denote $B_r$ to represent a random background image. As shown in Fig.~\ref{fig:main2}B, here $\tilde{I}_t = D_I(f_{2D}, f_{a}, B_r)$, in absence of access to a paired image to enforce a direct pixel-wise loss. Thus, the parameters of $E_P$ and $E_A$ are optimized for the unpaired samples using the following loss function, 
$\mathcal{L}_{\textit{\textit{UNP}}} = |p_{2D} - \tilde{p}_{2D}| + \lambda_2|f_a - \tilde{f}_a|$, where $\tilde{p}_{2D}=\mathcal{T}^{-1} \circ\mathcal{T}_k\circ E_P\circ\mathcal{T}(\tilde{I}_t)$ and $\tilde{f}_a=E_A(\tilde{I}_t)$. Here, $\mathcal{T}$ and $\mathcal{T}^{-1}$ represents a differentiable spatial transformation (such as image flip or in-plane rotation) and its inverse, respectively. We employ this to maintain a consistent representation across spatial-transformations. Note that, for the flip-operation of $p_{2D}$, we also exchange the indices of the joints associated with the left side to right and vice-versa.

We train on three different loss functions, viz. $\mathcal{L}_{prior}, \mathcal{L}_{P}$, and $\mathcal{L}_{\textit{UNP}}$ at separate iterations, each with different optimizer. Here, $\mathcal{L}_{prior}$ denotes the supervised loss directly on $p_{3D}$ and $p_{2D}$ for the synthetically rendered images on randomly selected backgrounds, as discussed before.

\begin{table*}[htp]
	\caption{ 
	Results on Human3.6M following the standard protocol-II setup. 
	Here, \textit{Sup.} (2nd column) denotes the amount of supervision accessed by the respective approaches. 
	Accordingly, the table is divided into 4 row-groups, \textbf{a)} row 1-5 use full 3D pose sup., \textbf{b)} row 6-10 use full 2D pose as weak sup. \textbf{c)} row 11-12: unsupervised approaches, and \textbf{d)} row 13: \textit{Ours(semi-sup.)}. We outperform prior approaches in both weakly supervised and unsupervised setting (highlighted as boldface).
	}
	\centering
	\setlength\tabcolsep{2.6pt}
	\resizebox{1.0\textwidth}{!}{
	\begin{tabular}{l|c|ccccccccccccccc|c}
	\hline
 		Protocol-II & Sup. & Direct. & Disc. & Eat & Greet & Phone & Photo & Pose & Purch. & Sit & SitD & Smoke & Wait & Walk & WalkD & WalkT & Avg.($\downarrow$) \\
		\hline\hline
		({\color{coolblack}Akhter et al.} \citeyear{akhter2015pose}) & Full-3D & 199.2 & 177.6 & 161.8 & 197.8 & 176.2 & 186.5 & 195.4 & 167.3 & 160.7 & 173.7 & 177.8 & 181.9 & 198.6 & 176.2 & 192.7 & 181.1\\
		\cite{zhou2016sparse} & Full-3D & 99.7 & 95.8 & 87.9 & 116.8 & 108.3 & 107.3 & 93.5 & 95.3 & 109.1 & 137.5 & 106.0 & 102.2 & 110.4 & 106.5 & 115.2 & 106.7 \\
		\cite{bogo2016keep} & Full-3D &  62.0 & 60.2 & 67.8 & 76.5 & 92.1 & 77.0 & 73.0 & 75.3 & 100.3 & 137.3 & 83.4 & 77.3 & 79.7 & 86.8 & 87.7 & 82.3\\
		({\color{coolblack}Moreno et al. }\citeyear{moreno20173d}) & Full-3D & 66.1 & 61.7 & 84.5 & 73.7 & 65.2 & 67.2 & 60.9 & 67.3 & 103.5 & 74.6 & 92.6 & 69.6 & 78.0 & 71.5 & 73.2 & 74.0 \\
		\cite{martinez2017simple} &Full-3D & 44.8 & 52.0 & 44.4 & 50.5 & 61.7 & 59.4 & 45.1 & 41.9 & 66.3 & 77.6 & 54.0 & 58.8 & 35.9 & 49.0 & 40.7 & 52.1 \\
		\hline

		\cite{wu2016single} & Full-2D & 78.6 & {90.8} & 92.5 & 89.4 & 108.9 & 112.4 & 77.1 & {106.7} & 127.4 & 139.0 & 103.4 & 91.4 & 79.1 & - & - & 98.4 \\
		\cite{tung2017adversarial} & Full-2D & {77.6} & 91.4 & {89.9} & {88.0} & {107.3} & {110.1} & {75.9} & 107.5 & {124.2} & {137.8} & {102.2} & {90.3} & {78.6} & - & - & {97.2} \\
		
		\cite{chen2019unsupervised} & Full-2D & - & - & - & - & - & - & - & - & - & - & - & - & - & - & - & 68.0 \\
		
		({\color{coolblack}Wandt et al.} \citeyear{wandt2019repnet}) & Full-2D &  53.0 & 58.3 & 59.6 & 66.5 & 72.8 & 71.0 & 56.7 &  69.6 &  78.3 &  95.2 &  66.6 &  58.5 &  63.2 &  57.5 &  49.9 &  65.1 \\
		
		Ours (weakly-sup.) & Full-2D & 56.0 & 53.2 & 56.3 & 63.6 & 74.1 & 77.5 & 53.4 & 67.9 & 75.8 & 90.8 & 64.2 & 56.9 & 61.4 & 56.3 & 49.7 & \textbf{63.8} \\
		
        \hline
        \rowcolor{gray!10}
		({\color{coolblack}Rhodin et al.} \citeyear{rhodin2018unsupervised}) & {Multi-view} & - & - & - & - & - & - & - & - & - & - & - & - & - & - & - & 98.2 \\ 
		
		\rowcolor{gray!10}
		Ours (unsup.) & \textbf{No sup.} & 80.2 & 81.3 & 86.0 & 86.7 & 94.1 & 83.4 & 87.5 & 84.2 & 101.2 & 110.9 & 86.0 & 87.8 & 86.9 & 94.3 & 90.9 & \textbf{89.4} \\ \hline
        
        \rowcolor{gray!25}
		Ours (semi-sup.) & 5\%-3D & 46.6 & 54.5 & 50.1 & 46.4 & 81.3 & 42.4 & 41.1 & 56.4 & 86.7 & 82.9 & 49.0 & 47.7 & 64.1 & 48.2 & 44.3 & \textbf{56.1} \\
		\hline
	\end{tabular}}
	\vspace{-4mm}
	\label{tab:protocol2results}
\end{table*} 

\begin{table}[htp]
	\footnotesize
	\caption{ 
	Results for the MPI-INF-3DHP dataset. 
	Here, \textit{Trainset} (2nd column) denotes access to 3DHP trainset images before evaluation. \textit{Sup.} (3rd column) denotes supervision level on 3DHP image-pose pairs. 4 row-groups, \textbf{a)} row 1-2: Fully supervised, \textbf{b)} row 3-7: Weakly supervised, \textbf{c)} row 8-10: Unsupervised, \textbf{d)} row 11: Semi-supervised.
	}
	\centering
	\setlength\tabcolsep{2.0pt}
	\resizebox{0.48\textwidth}{!}{
	\begin{tabular}{ll|cc|ccc}
	\hline
 		No.&Method & Trainset & Sup. & PCK ($\uparrow$) & AUC ($\uparrow$) & MPJPE ($\downarrow$) \\
		\hline\hline
		1.& \cite{mehta2017vnect} & +3DHP & Full-3D & 76.6 & 40.4 & 124.7 \\
		2.&({\color{coolblack}Rogez et al.} \citeyear{rogez2017lcr}) & +3DHP & Full-3D & 59.6 & 27.6 & 158.4 \\		
		
		\hline
		3.&\cite{zhou2017towards} & +3DHP & Full-2D & 69.2 & 32.5 & 137.1 \\
		
		4.&\cite{kanazawa2018end} & +3DHP & Full-2D & 77.1 &	40.7 & 113.2 \\
		5.&\cite{yang20183d} & +3DHP & Full-2D & 69.0 & 32.0 & - \\
		6.&\cite{chen2019unsupervised} & +3DHP & Full-2D &   {71.7}   & {36.3}  & - \\
		
		 
		7.&{Ours (weakly-sup.)} & +3DHP & Full-2D & \textbf{80.2} & \textbf{44.8} & \textbf{97.1} \\
		\hline
		
		\rowcolor{gray!10}
		8.&\cite{chen2019unsupervised} & -3DHP & - &   {64.3}   & {31.6}  & - \\
		\rowcolor{gray!10}
		9.&Ours (unsup.) & -3DHP & - & 76.5 & 39.8 & 115.3 \\ 
		\rowcolor{gray!10}
		10.&Ours (unsup.) & +3DHP & \textbf{No sup.} & \textbf{79.2} & \textbf{43.4} & \textbf{99.2} \\ \hline
		\rowcolor{gray!25}
		11.&Ours (semi-sup.) & +3DHP & 5\%-3D & \underline{81.9} & \underline{52.6} & \underline{89.8} \\
		\hline
	\end{tabular}}
	\vspace{-2mm}
	\label{tab:mpiinf3dhp}
\end{table} 


\section{Experiments}
In this section, we describe experimental details followed by a thorough analysis of the framework for bench-marking on two widely used datasets, Human3.6M and MPI-INF-3DHP.  


We use Resnet-50 (till \textit{res4f}) with ImageNet-pretrained parameters as the base pose encoder $E_P$, whereas the appearance encoder is designed separately using 10 Convolutions. $E_P$ later divides into two parallel branches of fully-connected layers dedicated for $v_k$ and $c$ respectively. We use $J=17$ for all our experiments as shown in Fig.~\ref{fig:main1}. The channel-wise aggregation of $f_{am}$ (16-channels) and $f_{hm}$ (17-channels) is passed through two convolutional layers to obtain $f_{2D}$ (128-maps), which is then concatenated with $f_a$ (512-maps) to form the input for $D_I$ (each with 14$\times$14 spatial dimension). Our experiments use different AdaGrad optimizers (learning rate: 0.001) for each individual loss components in alternate training iterations, thereby avoiding any hyper-parameter tuning. We perform several augmentations (color jittering, mirroring, and in-plane rotation) of the 20 synthetic samples, which are used to provide a direct supervised loss at the intermediate pose representations.

\textbf{Datasets.} The \textit{base-model} is trained on a mixture of two datasets, \ie Human3.6M and an in-house collection of YouTube videos (also refereed as YTube). In contrast to the in-studio H3.6M dataset, YTube contains human subjects in diverse apparel and BG scenes performing varied forms of motion (usually dance forms such as western, modern, contemporary etc.). Note that all samples from H3.6M contribute to the paired dataset $\mathcal{D}_p$, whereas $\sim$40\% samples in YTube contributed to $\mathcal{D}_p$ and rest to $\mathcal{D}_{unp}$ based on the associated BG motion criteria. However, as we do not have ground-truth 3D pose for the samples from YTube (in-the-wild dataset), we use MPI-INF-3DHP (also refereed as 3DHP) to quantitatively benchmark generalization of the proposed pose estimation framework.

\subsubsection{a) Evaluation on Human3.6M.}
We evaluate our framework on protocol-II, after performing scaling and rigid alignment of the poses inline with the prior arts~({\color{coolblack}Chen et al.~\citeyear{chen2019unsupervised}}; {\color{coolblack}Rhodin et al.~\citeyear{rhodin2018unsupervised}}). We train three different variants of the proposed framework \ie a) \textit{Ours(unsup.)}, b) \textit{Ours(semi-sup.)}, and c) \textit{Ours(weakly-sup.)} as reported in Table~\ref{tab:protocol2results}. After training the \textit{base-model} on the mixed YTube+H3.6M dataset, we finetune it on the static H3.6M dataset by employing $\mathcal{L}_{prior}$ and $\mathcal{L}_{p}$ (without using any multi-view or pose supervision) and denote this model as \textit{Ours(unsup.)}. This model is further trained with full supervision on the 2D pose landmarks simultaneously with $\mathcal{L}_{prior}$ and $\mathcal{L}_{p}$ to obtain \textit{Ours(weakly-sup.)}. Finally, we also train \textit{Ours(unsup.)} with supervision on 5\% 3D of the entire trainset simultaneously with $\mathcal{L}_{prior}$ and $\mathcal{L}_{p}$ (to avoid over-fitting) and denote it as \textit{Ours(semi-sup.)}. As shown in Table~\ref{tab:protocol2results}, \textit{Ours(unsup.)} clearly outperforms the prior-art~({\color{coolblack}Rhodin et al.~\citeyear{rhodin2018unsupervised}}) with a significant margin (89.4 vs. 98.2) even without leveraging multi-view supervision. Moreover, \textit{Ours(weakly-sup.)} demonstrates state-of-the-art performance against prior weakly supervised approaches. 

\begin{figure*}[!tbhp]
\begin{center}
	\includegraphics[width=1.00\linewidth]{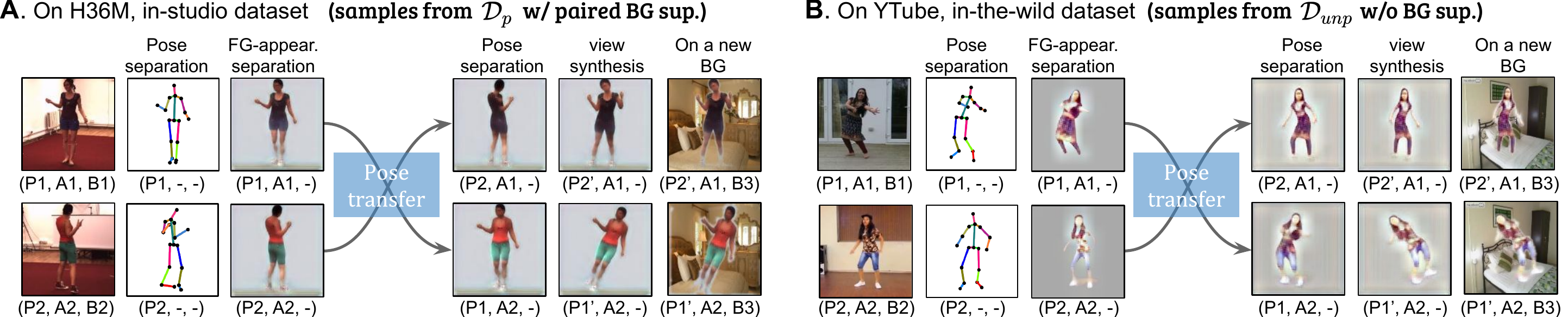}
	\vspace{-5.2mm}
	\caption{\small 
	Qualitative results, showing disentanglement of Pose (ID'd as P1 and P2), FG (ID'd as A1 and A2) and BG (ID'd as B1, B2, and B3). Images in first column (of each panel) define the IDs which are later used for novel image synthesis. 
	Devoid of a direct pixel-wise loss, energy-based losses for samples from $\mathcal{D}_{unp}$, help to clearly separate the FG person even in absence of a BG estimate (right panel).
	}
    \vspace{-2mm}
    \label{fig:viewsyn}  
\end{center}
\end{figure*}

\begin{figure*}[!tbhp]
\begin{center}
	\includegraphics[width=0.98\linewidth]{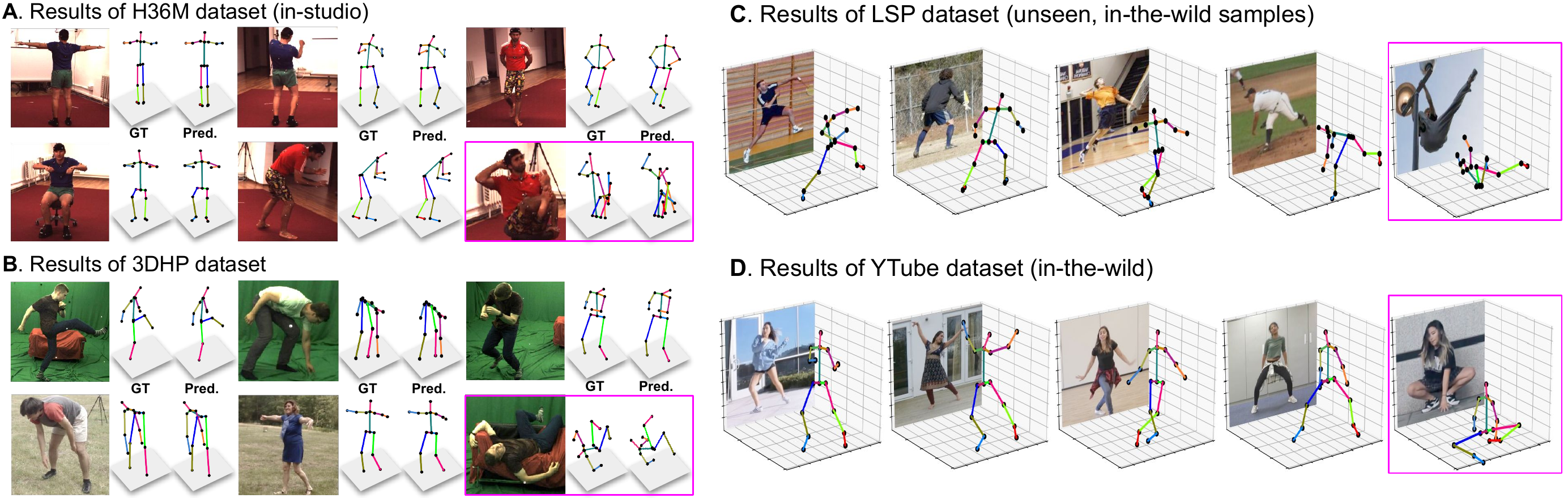}
	\vspace{-4mm}
	\caption{\small 
	Qualitative results on 4 different datasets. Note that, results on LSP is obtained in an unseen setting (\ie not even unpaired unsup. training). The pink box highlights some failure cases, specifically in presence of self-occlusion as a result of joint-position ambiguity.
	}
    \vspace{-2mm}
    \label{fig:qualitative}  
\end{center}
\end{figure*}

\subsubsection{b) Evaluation on MPI-INF-3DHP.}
We aim to realize a higher level of generalization in consequence of leveraging rich kinematic prior information. The proposed framework outputs 3D pose, which is bounded by the kinematic plausibility constraints even for unseen apparel, BG and action categories. This characteristic is clearly observed while evaluating performance of our framework on unseen 3DHP dataset. We take \textit{Ours(weakly-sup.)} model trained on YTube+H3.6M dataset to obtain 3D pose predictions on unseen 3DHP testset (9th row in Table~\ref{tab:mpiinf3dhp}). We clearly outperform the prior work~\cite{chen2019unsupervised} by a significant margin in a fully-unseen setting (8th and 9th row with -3DHP in Table~\ref{tab:mpiinf3dhp}). Furthermore, our weakly supervised model (with 100\% 2D pose supervision) achieves state-of-the-art performance against prior approaches at equal supervision level.

\begin{table}[t]
	\footnotesize
	\caption{ 
	Results on ablations of the proposed framework. It clearly highlights importance of $\mathcal{T}_{fk}$, $\mathcal{T}_m$, and use of $\mathcal{D}_{unp}$ in the unsupervised training pipeline. Notice the improvement in 3DPCK on the unseen 3DHP testset as a result of incorporating $\mathcal{D}_{unp}$ in the unsupervised training pipeline.
	}
	\centering
	\setlength\tabcolsep{6.0pt}
	\resizebox{0.46\textwidth}{!}{
	\begin{tabular}{l|c|cc}
	\hline
 		\multirow{2}{*}{\makecell{Method\\ (unsup.)}} & 
 		{Training set} & \multirow{2}{*}{\makecell{MPJPE on\\ H36M}} & \multirow{2}{*}{\makecell{3DPCK on\\ MPI-3DHP}} \\ 
 		& YTube+H3.6M \\
		\hline\hline
		Ours w/o $\mathcal{T}_{fk}$ & $\mathcal{D}_p$ & 134.8 & 47.9 \\
		Ours w/o $\mathcal{T}_m$ & $\mathcal{D}_p$ & 101.8 & 61.7 \\
		\hline
		Ours(unsup.) & $\mathcal{D}_p$ & 91.1 & 66.3 \\
		\rowcolor{gray!25}
		Ours(unsup.) & $\mathcal{D}_p \cup \mathcal{D}_{unp}$ & \textbf{89.4} & \textbf{71.2} \\
		\hline
	\end{tabular}}
	\vspace{-2mm}
	\label{tab:ablations}
\end{table} 

\subsubsection{c) Ablation study.}
In the proposed framework, our major contribution is attributed to the design of differentiable transformations and an innovative way to facilitate the usage of unpaired samples even in presence of BG motion. Though effectiveness of camera-projection has been studied in certain prior works~\cite{chen2019unsupervised}, use of forward-kinematic transformation $\mathcal{T}_{fk}$ and affinity map in the spatial-map transformation $\mathcal{T}_m$ is employed for the first time in such a learning framework. Therefore, we evaluate importance of both $\mathcal{T}_{fk}$ and $\mathcal{T}_m$ by separately bypassing these modules through neural network transformations. Results in Table~\ref{tab:ablations} clearly highlight effectiveness of these carefully designed transformations for the unsupervised 3D pose estimation task.

\subsubsection{d) Qualitative results.} Fig.~\ref{fig:viewsyn} depicts qualitative results derived from \textit{Ours(unsup.)} on in-studio H3.6M and in-the-wild YTube dataset. It highlights effectiveness of unsupervised disentanglement through separation or cross-transfer of apparel, pose, camera-view and BG, for novel image synthesis. Though, our focus is to disentangle 3D pose information, separation of apparel and pose transfer is achieved as a byproduct of the proposed learning framework. In Fig.~\ref{fig:qualitative} we show results on the 3D pose estimation task obtained from \textit{Ours(weakly-sup.)} model. Though we train our model on H3.6M, 3DHP and YTube datasets, results on LSP dataset~\cite{johnson2010clustered} is obtained without training on the corresponding train-set, \ie in a fully-unseen setting. Reliable pose estimation on such diverse unseen images highlights generalization of the learned representations thereby overcoming the problem of dataset-bias.

\section{Conclusion}


We present an unsupervised 3D human pose estimation framework, which relies on a minimal set of prior knowledge regarding the underlying kinematic 3D structure.
The proposed local-kinematic model indirectly endorses a kinematic plausibility bound on the predicted poses, thereby limiting the model from delivering implausible pose outcomes. 
Furthermore, our framework is capable of leveraging knowledge from video frames even in presence of background motion, thus yielding superior generalization to unseen environments. 
In future, we would like to extend such frameworks for predicting 3D mesh, by characterizing the prior knowledge on human shape, alongside pose and appearance.



{
\noindent
\textbf{Acknowledgements.} This work was supported by a Wipro PhD Fellowship (Jogendra) and in part by DST, Govt. of India (DST/INT/UK/P-179/2017).}

{\small
  \bibliographystyle{aaai}
  \bibliography{ms}
}

\end{document}